\documentclass[conference]{IEEEtran}

\usepackage{graphicx}
\usepackage{subfig}
\usepackage{multirow}
\usepackage{array}
\usepackage{booktabs}
\usepackage{fancyhdr}
\fancypagestyle{empty}{%
  \fancyhf{}
  \fancyhead[L]{This is a pre-print version to appear in IEEE ICMLA 2016.}
}

\usepackage[colorlinks]{hyperref}

\renewcommand\cap[3]{\caption[#2]{\label{#1}\textsc{#2}. \small\textit{#3}}}

\title{Assessing Threat of Adversarial Examples \\ on Deep Neural Networks}

\author{\IEEEauthorblockN{Abigail Graese, Andras Rozsa, and Terrance E. Boult }
\IEEEauthorblockA{University of Colorado at Colorado Springs\\
Vision and Security Technology (VAST) Lab\\
Email: http://vast.uccs.edu/contact-us
}}

\begin{document}

\maketitle

\thispagestyle{empty}

\begin{abstract}

Deep neural networks are facing a potential security threat from adversarial examples, inputs that look normal but cause an incorrect classification by the deep neural network.  For example, the proposed threat could result in hand-written digits on a scanned check being incorrectly classified but looking normal when humans see them.  This research assesses the extent to which adversarial examples pose a security threat, when one considers the normal image acquisition process.  This process is mimicked by simulating the transformations that normally occur in acquiring the image in a real world application, such as using a scanner to acquire digits for a check amount or using a camera  in an autonomous car.  These small transformations negate the effect of the carefully crafted perturbations of adversarial examples, resulting in a correct classification by the deep neural network. Thus just acquiring the image decreases the potential impact of the proposed security threat.  We also show that the already widely used process of averaging over multiple crops neutralizes most adversarial examples. Normal preprocessing, such as text binarization, almost completely neutralizes adversarial examples.  This is the first paper to show that for text driven classification, adversarial examples are an academic curiosity, not a security threat.

\end{abstract}

\hfill


\section{Introduction}

The solving of classification problems in machine learning has recently made significant progress through the use of deep neural networks, or deep learning~\cite{c7, c8, c11}.  Deep convolutional neural networks (DCNNs) can be used for supervised learning, which is the use of a training set that contains known outputs for the inputs during the training of the DCNN.  After training is completed, when presented with unknown inputs, the DCNN is able to classify the inputs with exceptional accuracy.

Although DCNNs have a high accuracy rate, images known as adversarial examples trick DCNNs into classifying an image incorrectly despite humans seeing almost no difference between the original and adversarial image.  Recently, researchers have proposed that adversarial examples can ``seriously undermine the security of the system supported by the DCNN''~\cite{c3}, because the incorrect classification could potentially lead to an incorrect action with consequences.  For example, if a stop sign was crafted as an adversarial example, an autonomous vehicle could complete an incorrect classification of the sign and cause an accident to occur~\cite{c12}. If the digits signifying an amount on a check were crafted to be adversarial, the amount of money transferred between accounts could be altered.

In real world applications of deep learning which use inputs from a camera or scanned images, such as an autonomous car or processing of checks as mentioned above, the input images will never be perfectly captured.  The input images will always contain slight transformations, such as shifting or blurring, and perturbations, such as noise, due to the imperfect capture of the input, which slightly perturbs the input to the neural network from the intended input.  As we will see, these slight transformations render the majority of adversarial examples nonadversarial, thus, no additional defensive technique is necessary.  In this paper, the acquisition process is mimicked by performing small transformations that could be expected in normal image acquisition.

The validity of a situation where an adversarial example could be crafted into a real world application input and then survives the image acquisition process needs to be assessed. This research assesses the extent to which adversarial examples are handled and classified correctly, simply through the natural process of acquiring the image, which renders the input nonadversarial.

\begin{figure}
\vspace{0.05cm}
\centering
\includegraphics[width=.98\columnwidth]{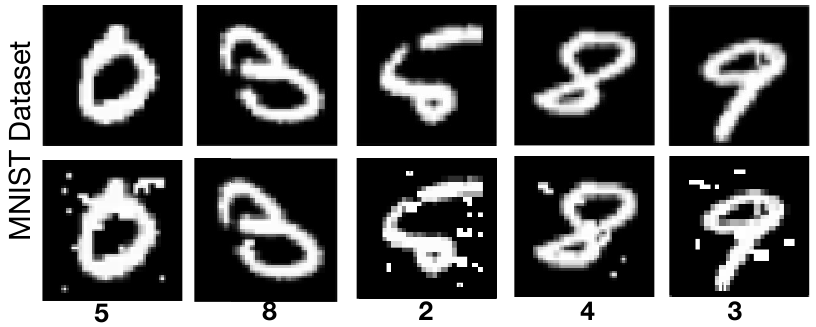}
\cap{fig:examples}{Authentic Versus Adversarial Examples} {The images on the top row of the figure are legitimate images.  The images on the bottom row of the figure are adversarial examples, and the numbers below each of those images are the numbers that the DCNN mistakenly classifies the adversarial examples.  Adapted from~\cite{c3}.}
\vspace{-0.05cm}

\end{figure}

\begin{figure*}[t]
\small
\begin{center}
	\vspace{-0.25cm}
	\begin{tabular}{cccccc}
		\multirow{-2}[1]{*}{
		\subfloat[][\label{fgs_fgv:a}\centering Original]{\includegraphics[width=1.5cm]{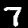}}}	 		\hspace{0.6cm}
	  	& \subfloat[][\label{fgs_fgv:b}\centering FGS \par 0.4860\,$\vert$\,549\,$\vert$\,26]{\makebox[1.3\width]{\includegraphics[width=1.5cm]{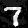}}}
 		\hspace{0.6cm}
	  	& \subfloat[][\label{fgs_fgv:c}\centering FGS \par 0.3885\,$\vert$\,1070\,$\vert$\,51]{\makebox[1.3\width]{\includegraphics[width=1.5cm]{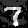}}}
 		\hspace{0.6cm}
	  	& \subfloat[][\label{fgs_fgv:d}\centering FGS \par 0.3478\,$\vert$\,1334\,$\vert$\,64]{\makebox[1.3\width]{\includegraphics[width=1.5cm]{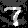}}}
 		\hspace{0.6cm}
	  	& \subfloat[][\label{fgs_fgv:e}\centering FGS \par 0.3132\,$\vert$\,1578\,$\vert$\,76]{\makebox[1.3\width]{\includegraphics[width=1.5cm]{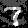}}}
 		\hspace{0.6cm}
	  	& \subfloat[][\label{fgs_fgv:f}\centering FGS \par 0.1949\,$\vert$\,2604\,$\vert$\,127]{\makebox[1.3\width]{\includegraphics[width=1.5cm]{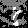}}}
		\vspace{-0.15cm}
	  \\
		& \subfloat[][\label{fgs_fgv:g}\centering FGV \par 0.7727\,$\vert$\,358\,$\vert$\,57]{\makebox[1.3\width]{\includegraphics[width=1.5cm]{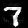}}}
 		\hspace{0.6cm}
		& \subfloat[][\label{fgs_fgv:h}\centering FGV \par 0.6527\,$\vert$\,666\,$\vert$\,113]{\makebox[1.3\width]{\includegraphics[width=1.5cm]{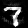}}}
 		\hspace{0.6cm}
		& \subfloat[][\label{fgs_fgv:i}\centering FGV \par 0.5665\,$\vert$\,969\,$\vert$\,170]{\makebox[1.3\width]{\includegraphics[width=1.5cm]{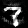}}}
 		\hspace{0.6cm}
		& \subfloat[][\label{fgs_fgv:j}\centering FGV \par 0.4919\,$\vert$\,1253\,$\vert$\,226]{\makebox[1.3\width]{\includegraphics[width=1.5cm]{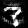}}}
 		\hspace{0.6cm}
		& \subfloat[][\label{fgs_fgv:k}\centering FGV \par 0.4258\,$\vert$\,1514\,$\vert$\,252]{\makebox[1.3\width]{\includegraphics[width=1.5cm]{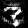}}}
		\vspace{-0.15cm}
	\end{tabular}
\end{center}

\cap{fig:fgs_fgv}{FGS Versus FGV Adversarial Examples}{The metrics underneath the numbers are the PASS, \(L_2\) norm and \(L_\infty\) norm respectively.  Image (a) is the original MNIST image.  Images (b)-(f) are FGS adversarial examples.  Image (b) has the minimum perturbation and \(\epsilon\) required to create an adversarial example.  Image (c), (d), (e) and (f) have an \(\epsilon\) of 0.20, 0.25, 0.30, and 0.50 respectively.  For MNIST, the pixels have basically binary values  (either 0 or 255) and applying an \(\epsilon\) of 0.20 with FGS forms perturbations with $L_\infty$\,=\,51.  Human perception can see deviations of 5-10 gray values when doing a comparison.  Images (g)-(k) are FGV adversarial examples.  Image (g) has the minimum perturbation required to create an adversarial.  Image (h)-(k) have 2, 3, 4, and 5 times the minimum perturbations respectively. }
\end{figure*}

We also note that most state-of-the-art deep convolutional neural networks (DCNNs) use multiple crops and often multiple networks in reaching their final decision.  To date, no paper on adversarial examples has examined whether they could survive this widely used component of DCNNs.  We show that even with only 5 crops (on non-transformed), compared to the 10s to hundreds used in state of the art networks,  the majority of adversarial images will be correctly classified.  \par
The problem addressed in this paper is whether adversarial examples are a security threat to DCNNs and the extent to which this is or is not the case.

\section{Related Work}

Deep neural networks are learning models that produce state-of-the-art results for several types of classification and recognition problems~\cite{c7,c8}. Szegedy et al.~\cite{c4} discovered that there exist perturbations which when included in an image cause an incorrect classification by the DCNN but they are ``imperceptible'' to humans.  These examples were named ``adversarial examples.'' 

Since this discovery, several advancements in the understanding and creation of adversarial examples have taken place. Sabour et al.~\cite{c16} demonstrated that the existence of adversarial examples could be the result of the architecture of DCNNs themselves.  Goodfellow et al.~\cite{c5} presented the fast gradient sign (FGS) method for generating adversarial examples, which forms perturbations $\eta_{_{fgs}}$ using the ``sign of the elements of the gradient of the cost function with respect to the input,'' that is defined as
\begin{equation}
	\label{eq:fgs}
	\eta_{_{fgs}} = \epsilon\,sign(\nabla_xJ(\theta,x,y))
\end{equation}
where \(x\) is the input to the model, \(y\) is the target of \(x\), and \(J(\theta,x,y)\) is the cost used to train the network.\par

Rozsa et al.~\cite{c1} extended upon the FGS approach for generating adversarial examples and introduced two effective ways to produce more robust adversarial images using the fast gradient value (FGV) and the hot/cold (HC) approaches. The fast gradient value (FGV) approach uses ``a scaled version of the raw gradient of loss'' to create adversarial examples with distortions even less perceptible to humans.  The perturbation formed by the FGV approach can be defined as
\begin{equation}
	\label{eq:fgv}
	\eta_{_{fgv}}= \epsilon\,\nabla_xJ(\theta,x,y).
\end{equation}


The hot/cold approach defines a hot class as the target classification class and a cold class as the original classification class.  This method then uses these defined classes to create features that cause classification to move towards the hot or target class from the original cold class.  In addition to the different approaches to generating adversarial examples, Rozsa et al. also defined a metric for quantifying adversarial examples by measuring both the element-wise difference and probability that the image could be a different perspective of the original input called a Perceptual Adversarial Similarity Score (PASS).  This score is a number between 0 and 1, where 1 denotes an adversarial example with no visible difference from the original image.\par 
Rozsa et al.~\cite{c1} also explored the effectiveness of fine-tuning a DCNN with adversarial examples and showed that such networks were able to correctly classify 86\% of previously adversarial examples, demonstrating the ability to increase the robustness of a DCNN to adversarial examples with the right training.  In the context of this paper, the results attained by fine-tuning with adversarial examples are complementary to the results shown below.

Rozsa et al.~\cite{c2} also contributed to the known information about adversarial images by defining adversarial examples that exist in nature as ``an image that is misclassified, but that will be correctly classified when an imperceptible modification is applied.''  Natural adversarial images demonstrate an additional aspect of the security threat of adversarial examples that needs to be considered.  

In response to the growing interest and research in adversarial examples, Papernot et al.~\cite{c3} asserted that by using deep learning algorithms, system designers made security assumptions about DCNNs, specifically in reference to adversarial samples.   Papernot et al.~\cite{c3} attempted to address the problem by demonstrating the use of distillation as an alternative form of training a DCNN to increase the percentage of correctly classified adversarial examples when presented to the DCNN as inputs using the CIFAR-10~\cite{c11} and MNIST~\cite{c9} data sets.  However, Carlini et al.~\cite{c23} demonstrated that the assumptions made in this approach are wrong and the approach itself is not effective.  

Papernot et al.~\cite{c12} also presented a method for attacking a DCNN with adversarial examples without prior knowledge of the architecture of the network itself, and only having access to the targeted network's output and some knowledge of the type of input.  To accomplish this, Papernot et al. trained a substitute DCNN on possible inputs for the targeted DCNN.  After the network was trained, adversarial examples were crafted with the FGS method~\cite{c5}.  These examples were generated with different values of \(\epsilon\), as defined in Eq.~\ref{eq:fgs}.  Examples of the FGS adversarial images generated with various values of \(\epsilon\) can be seen in Fig.~\ref{fig:fgs_fgv}.  The examples generated with higher values of \(\epsilon\) do not fit the portion of the definition of adversarial examples that the perturbations need to be imperceptible to human observers.

A different technique for increasing a DCNN's ability to handle and correctly classify adversarial examples was put forward by Luo et al.~\cite{c15}.  This technique uses a ``transformation of the image that selects a region in which the convolutional neural network (CNN) is applied, denoted a foveation, discarding the information from the other regions'' as the input to the DCNN.  This technique enabled the DCNN to correctly classify over 70\% of adversarial examples.

The previous state of the art~\cite{c18} takes crops of the original input and uses the average prediction over the classified crops which mimics natural perturbations.  Crops like the ones used in GoogLeNet predate the discovery of adversarial examples and are used to improve accuracy of DCNNs.  The improvement of accuracy in the DCNN also also helped to mitigate classification errors caused by adversarial examples.  By taking crops of images, the accuracy of the DCNN when classifying adversarial examples is greatly increased.  The networks tested by Papernot et al.~\cite{c3} did not use this previous state-of-the-art.  This research aims to assess the level of threat imposed by adversarial examples to DCNNs as suggested by Papernot et al. ~\cite{c3}. 

\section{Method}
 The proposed security threat provided that in a critical situation, the incorrect classification of an adversarial example made by the DCNN could cause actions with immense repercussions.
 
The image acquisition process as described above, which is necessary in physical all real world applications of classification by a DCNN, is always going to capture an imperfect input.

\begin{figure}[t]{
  \small
  \setlength{\tabcolsep}{1pt}
  \newcolumntype{V}{ >{\centering\arraybackslash}m{1cm}}
  \begin{tabular}[h]{VVVVVVVV}
    & a & b & c & d & e & f & g \\
    Clean &
    {\includegraphics[width=1cm]{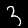}}&
    {\includegraphics[width=1cm]{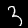}}&    
    {\includegraphics[width=1cm]{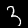}}&    
    {\includegraphics[width=1cm]{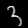}}&    
    {\includegraphics[width=1cm]{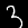}}&    
    {\includegraphics[width=1cm]{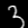}}&    
    {\includegraphics[width=1cm]{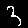}}\\
    FGS &
    {\includegraphics[width=1cm]{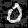}}&
    {\includegraphics[width=1cm]{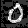}}&    
    {\includegraphics[width=1cm]{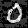}}&    
    {\includegraphics[width=1cm]{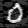}}&    
    {\includegraphics[width=1cm]{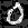}}&    
    {\includegraphics[width=1cm]{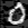}}&    
    {\includegraphics[width=1cm]{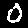}}\\
    FGV &
    {\includegraphics[width=1cm]{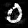}}&
    {\includegraphics[width=1cm]{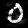}}&    
    {\includegraphics[width=1cm]{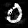}}&    
    {\includegraphics[width=1cm]{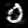}}&    
    {\includegraphics[width=1cm]{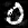}}&    
    {\includegraphics[width=1cm]{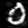}}&    
    {\includegraphics[width=1cm]{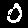}}\\
\end{tabular}

\cap{fig:trans}{Image Transformations}{The rows show transformations of a clean image, FGS, and FGV adversarial samples. Column (a) is the original image.  Column (b) has had one column translated to black.  Column (c) has a small amount of noise added to it.  Column (d) has the blur kernel of (2,1) applied to it.  Column (e) has been cropped to 27\,$\times$\,27 pixels and then resized back to 28\,$\times$\,28 pixels. Column (f) is a combination of all previously mentioned transformations. Column (g) is the result of binarization with Oshu thresholding.}}

\end{figure}

To assess the extent of the potential security threat that adversarial examples cause for DCNNs after acquisition, it is proposed that a slight transformation -- such as a blur or a shift that would occur in normal image acquisition -- to be applied to images before they are classified by the DCNN.  To evaluate the effectiveness of this method, a trained deep convolutional neural network, a dataset including adversarial examples, and transformations are used.
\subsection{The Deep Neural Network}
A LeNet~\cite{c13} deep convolutional neural network, trained by the authors of~\cite{c1}, is used for completing classification experiments on the chosen dataset.  As listed in Tab.~\ref{table:mnist}, this network classifies images of the MNIST test set with an accuracy of 98.96\%.  
\subsection{The Dataset}
The DCNN is trained on the MNIST dataset of handwritten digits~\cite{c9}, and is tested with the MNIST test set.  This dataset provides a basis for a possible security weakness of the DCNN to adversarial examples.  If adversarial examples cause an incorrect classification of a handwritten selection, such as an amount on a check, it could cause the amount to be misinterpreted and cause an incorrect amount to be withdrawn.\par  
In addition to the MNIST test set~\cite{c9}, adversarial examples generated using the techniques in~\cite{c1} and ~\cite{c5} are also tested to initially demonstrate the effectiveness of adversarial examples, and then to assess the effectiveness of the transformations, which mimic the acquisition process, at negating the effect of the perturbations that create an adversarial example.  \par
In initial acquisition, MNIST images were subject to a pipeline of downsampling to 20\,$\times$\,20 pixels, binarization, and subsequent upsampling to 28\,$\times$\,28 pixels.
The adversarial examples used in the following experiments were generated by the authors of~\cite{c1} for the network described previously.  When generating FGS adversarial examples, the authors of~\cite{c1} increased \(\epsilon\), as defined in Eq.~\ref{eq:fgs}, until the image was made adversarial, so the entire data set is originally adversarial for the network described above.

\subsection{Transformations}
The aim of this research is to assess the extent to which adversarial examples can be handled and classified correctly, simply by the imperfection of the natural image acquisition process -- i.e., slight transformations and perturbations added to the input images for classification -- and understanding the impact each of these transformations has on classification. In order to mimic the image acquisition process as accurately as possible, an image was printed out, scanned back in, and analyzed for the types of transformation needed to closely replicate the effect of the acquisition. The types of transformations and perturbations that have been used to complete an experiment and their justifications are listed below.

\subsubsection{Translation}
The addition of a translation to the images processed by the DCNN replicates an alignment issue that could occur in the normal image acquisition process.  This was implemented by shifting the image to the right by one pixel and filling the pixels in the empty column with values of 0, which in RGB values is black.

\subsubsection{Noise}
In the image acquisition process, it is normal to see additive noise on an image, such as black dots seen when an image is scanned in.  To replicate the additive noise on an image, a small amount of generated noise is added to an input image before allowing the image to be processed and classified by the DCNN.  The noise mask was generated with a standard deviation of 0.25, and a mean of 0.  The noise mask was then added to a copy of the original image.

\subsubsection{Blurring} 
When acquiring an image in a real world environment, such as with a camera or scanner, it is virtually impossible to capture an image without any blurring. To replicate this, the amount of blurring was estimated from the printed image.  This led to the application of a blur kernel of (2,1) to the input images, as there was approximately one pixel of blur in the x direction and one-half pixel of blur in the y direction on the acquired image, with the asymmetry probably due to the scanner having a moving linear sensor.  

\subsubsection{Cropping \& Resizing}
This transformation mimics the event when the acquired image is smaller than the original image.  In order to mimic this, input images were cropped, the cropped image was saved, and then the cropped image was resized (using a cubic interpolation function) back to the image size expected by the DCNN of 28\,$\times$\,28 pixels.

\subsubsection{Combination}
The above transformations each demonstrate pieces of the whole image acquisition process.  In order to fully synthetically capture this process, the described transformations must all be applied to the input images.  Transformations were applied in the following order:  translation, noise, blur, crop and resize. This order was chosen to mimic the order in which the transformations occur in the natural image acquisition process.  After transformations were applied, the transformed image was input to the DCNN for classification.

\subsection{Fine-Tuning}
The described experiments were run both on the raw LeNet~\cite{c13} network, and on a fine-tuned network.  The network was fine-tuned because when it was trained, the network learned from clean images without transformations. When inputting transformed images for classification, the network was not robust enough to correctly classify transformed inputs -- even if the inputs were not adversarial examples -- with the same level of accuracy as with clean images.\par
In order to fine-tune the network, a set of 100k images was taken for training and 20k images were used for validation.  The dataset used for fine-tuning contained a total of 60k clean images and 60k transformed images, where the transformed images were the MNIST training set images with the combination of all transformations applied.  The fine-tuned network has an accuracy of 99.35\% on the chosen validation set, and 99.09\% on the MNIST test set.

\subsection{Fusion of Crops}
In order to more accurately mimic the previous state-of-the-art deep neural networks~\cite{c18}, a series of crops was implemented.  As was previously mentioned, crops are used to increase the accuracy of DCNNs.  With only a 28\,$\times$\,28 image for MNIST, the number and size of crops is more limited, so we used only 5 crops:  a center crop and 4 corner crops of size 27\,$\times$\,27 pixels rescaled back to 28\,$\times$\,28 pixels.  Each image resizing used a cubic interpolation function.  Each crop was classified by the DCNN separately, and their predictions were summed up to determine the predicted label.

\subsection{Binarization}
As is common in hand-written text recognition~\cite{c19},  before applying the recognition engine, the image is subject to preprocessing including binarization and noise removal. As mentioned above, the MNIST dataset~\cite{c9} was subject to such preprocessing before being compiled into the dataset used in training and experimentation.  The exact preprocessing of the MNIST images cannot be exactly replicated, due to an unclear description including lack of details on how down-sampling, binarization and subsequent up-sampling were performed.  Without details of the rescaling steps,  we approximate what we consider the most important step, binarization,  using an OpenCV~\cite{c20} version  of OTSU thresholding~\cite{c21}.  Binarization  is a critical step and takes into account the fact that machines are trained on basically binary data.  When it is forced to deal with data which is not binary, the machine is more easily confused. As we shall see, this single assumption may account for almost all the effectiveness of adversarial examples, and proper preprocessing renders them neutralized. Pure binarization may not be effective on the type of noise in~\cite{c3}, but the despeckeling/noise removal that is commonly used for document processing ~\cite{c19,c22}, would likely remove most of that noise as well.


\begin{table}[t]
\footnotesize
\centering\vspace*{1.55ex}
\begin{tabular}{r|c|c}
\toprule
TRANSFORMATION              				& RAW MODEL	& FINE-TUNED\\ \midrule \midrule
None                        				& 98.96\%	& 99.09\%\\
Translation of one column   				& 94.95\%	& 99.17\%\\ 
Noise                       				& 98.95\%	& 99.09\%\\ 
Blur   	                    				& 98.70\%	& 99.14\%\\ 
Crop (27\,$\times$\,27 px) and Resize	& 98.35\%	& 99.14\%\\ 
Combination								& 97.66\% 	& 98.88\%\\ \midrule 
5 crops (on non-transformed)   			& 98.67\%	& 99.12\%\\
Binarize (on non-transformed)   			& 98.76\%	& 99.04\%\\
\bottomrule
\end{tabular}%
\cap{table:mnist}{Accuracy on MNIST Test Set}{This table reports the results of our experiments on the MNIST test set. The column on the right shows the accuracy of the DCNN that was fine-tuned on transformed images. For comparison, we also show accuracies obtained on a raw model in the middle that was trained regularly. Accuracy is the proportion of correctly classified images that were transformed before testing as specified on the left.}
\end{table}

\begin{table}[t]
\footnotesize
\centering
\begin{tabular}{r|c|c}
\toprule
TRANSFORMATION              				& RAW MODEL	& FINE-TUNED\\ \midrule \midrule
None                        				& 00.00\%  	& 56.93\%\\
Translation of one column   				& 65.29\%	& 68.93\%\\
Noise                       				& 28.41\%	& 59.84\%\\
Blur                        				& 58.60\%  	& 59.83\%\\
Crop (27\,$\times$\,27 px) and Resize  	& 78.28\%	& 80.01\%\\
Combination								& 79.68\%	& 83.98\%\\ \midrule 
5 crops (on non-transformed)    			& 90.94\% 	& 81.66\%\\
Binarize (on non-transformed)   			& 99.24\% 	& 99.21\%\\
\bottomrule
\end{tabular}
\cap{table:fgs}{Accuracy on FGS Samples}{This table reports the results of the conducted experiments on a set of 10k FGS~\cite{c5} adversarial examples. We show accuracies obtained by testing on a regularly trained ``raw'' model (in the middle) and a fine-tuned model that was trained on additional transformed images as well. Accuracy is the proportion of correctly classified FGS adversarial images that were transformed before testing as specified on the left.}
\end{table}

\begin{table}[t]
\footnotesize
\centering\vspace*{1.55ex}
\begin{tabular}{r|c|c}
\toprule
TRANSFORMATION              				& RAW MODEL	& FINE-TUNED\\ \midrule \midrule
None                        				& 00.00\%    & 62.14\%\\
Translation of one column   				& 68.26\%	& 73.15\%\\
Noise                       				& 57.84\%  	& 64.59\%\\
Blur                        				& 64.77\% 	& 65.54\%\\ 
Crop (27\,$\times$\,27 px) and Resize 	& 76.16\%  	& 78.70\%\\
Combination								& 71.29\%  	& 75.95\%\\ \midrule 
5 crops (on non-transformed)   			& 81.66\%  	& 76.69\%\\
Binarize (on non-transformed)   			& 99.24\%  	& 98.88\%\\
\bottomrule
\end{tabular}%
\cap{table:fgv}{Accuracy on FGV Samples}{This table reports the results of the conducted experiments on a set of 10k FGV~\cite{c1} adversarial examples. We show accuracies obtained on a regularly trained ``raw'' model (in the middle) and a fine-tuned model that was trained on additional transformed images as well. Accuracy is the proportion of correctly classified FGV adversarial images that were transformed before testing as specified on the left.}
\end{table}


\section{Experiments \& Results}
\subsection{Procedure}
The experimentation procedure involved three datasets:  the MNIST test set~\cite{c9} of 10k images, a set of 10k randomly chosen FGS~\cite{c5} adversarial examples, and a set of 10k randomly chosen FGV~\cite{c1} adversarial examples.  All experiments were run on all three datasets in order to generate a basis of comparison for results.  After running a baseline with no transformations, experiments consisted of applying a transformation or combination of transformations to input images and then passing the transformed inputs to the DCNN for classification.  The results of the classification were measured by the percentage of images correctly classified by the DCNN.

\subsection{Transformation Results}
The specific results of the experiments completed are detailed in Tab.~\ref{table:mnist},~\ref{table:fgs}, and~\ref{table:fgv}.  These results demonstrate that the image acquisition process allows the DCNN to correctly classify a large portion of what used to be adversarial examples.\par

The most effective transformation to allow the DCNN to compute a correct classification is the cropping and subsequent resizing of  input images, which demonstrated 78.28\% and 76.16\% accuracy on the FGS~\cite{c5} and FGV~\cite{c1} datasets respectively for the raw network.  This transformation also only slightly altered the accuracy of the raw DCNN on the MNIST test set~\cite{c9} (from 98.96\% to 98.35\%).  Cropping and resizing also led to the highest accuracy on the fine-tuned network, demonstrating 80.01\% and 78.70\% accuracy on the FGS~\cite{c5} and FGV~\cite{c1} datasets respectively and 99.14\% on the MNIST test set.  After fine-tuning the network, the classification accuracy of adversarial examples increased, 56.93\% and 62.09\% on the FGS~\cite{c5} and FGV~\cite{c1} datasets respectively without applying transformations.\par

The results of this  portion of the research demonstrates that in the majority of cases, the effect of perturbations added to make FGV adversarial examples are more easily negated than the perturbations formed by the FGS method.  Intuitively, this is the case because the perturbations of FGS adversarial examples are bigger and more noticeable, and are therefore more likely to survive the transformations demonstrated in the natural image acquisition process.\par 

It should be noted that when transformations are applied, the performance of the deep neural network decreases on the MNIST test set.  This decrease is the result of the added transformations essentially creating natural adversarial examples, as defined in~\cite{c2}.  These natural adversarial examples introduce a different problem to the DCNN, because when put in the same situation where an incorrect classification causes an incorrect action, the natural adversarial examples would also cause an incorrect action.  While these actions would not be chosen by an adversary, there would still be actions with consequences.\par
Although testing of the other transformations and the combination of transformations have not produced results where the transformation is completely negating the effect of the adversarial examples, all of the transformations have improved upon the accuracy rate of the DCNN on adversarial examples without transformations.\par

\subsection{Fusion of Crops Results}
Doing experiments with the application and fusion of 5 crops was motivated by using the technique of the previous state of the art~\cite{c18} to improve accuracy of the DCNN.  These experiments produced the second highest number of correctly classified adversarial examples out of all of the experiments.

\subsection{Binarization Results}
Binarization produced the best results out of the transformations, achieving close to the performance of the deep neural network on the MNIST test set without any transformations.  In the binarization process, more FGS adversarial examples are correctly classified than FGV adversarial examples and thus are not surviving the image acquisition process.  This is due to the fact that FGS adversarial examples depend on bigger and brighter collections of noise to render the image adversarial in comparison to FGV adversarial examples.  

\subsection{Adversarial Examples on ImageNet}
After seeing the modest success of the synthetic image acquisition process on handling adversarial examples, an initial experiment was run on a GoogLeNet deep neural network~\cite{c18} on a subset of the ImageNet dataset~\cite{c17}, with 15k FGS adversarial examples, all of which were provided by the authors of~\cite{c1}.  The experiment consisted of applying the combination of transformations, as is described above.  The results of this experiments demonstrated that 63\% of adversarial examples were classified correctly for top-1 accuracy and 89.95\% of adversarial examples were classified correctly for top-5 accuracy.  The fact that the application of transformations are producing similar results for the MNIST dataset and a portion of the ImageNet dataset demonstrates that foveation as described in~\cite{c15} could be applied to any place in the image to negate the effect of adversarial examples.

\section{Conclusion \& Discussion}
 
This project assessed the extent to which adversarial examples are a security threat and demonstrated the effectiveness of simple solutions, such as slight transformations to the inputs, at mitigating that threat.  This research demonstrated that slight transformations do render the majority of FGS and FGV adversarial examples as nonadversarial.  The best results of this research, achieved through binarization of the inputs to the DCNN, demonstrated performance near to that of the deep neural network on clean images.  For comparison, this method classified 20\% more images correctly than the foveation method.  These results presented that for the MNIST data set~\cite{c9}, the potential security problem is negligible, as the adversarial examples can be almost completely mitigated through binarization, which is part of the acquisition process of the original images.  Therefore, when considering the application of digit classification on checks, adversarial examples are not a problem.\par  
Outside of the classification of handwritten digits, when considering an autonomous car, the camera capturing input for the DCNN has the opportunity to capture a traffic sign hundreds of times, each at a slightly different angle, rotation, alignment and blur.  This makes the chances of an adversary producing an adversarial example that would survive the image acquisition process significantly smaller than suggested by \cite{c23}.  If, independently, each frame that consists of an adversarial example is correctly classified 90\% of the time, then to get a majority wrong -- say 15 in 30 frames (1 second) -- would only have a ${30 \choose 15}(0.1)^{15} \approx 1.55\,\times\,10^{-6}$, i.e. about 1 in a million chance of causing an error.  \par
However, further research should focus on the effect of the natural image acquisition process on adversarial examples on a dataset such as ImageNet~\cite{c17} in order to formalize and assess the extent of the possible security threat to deep neural networks in real world applications.\par

It should also be noted, that this research does not take into account the case where an input from an adversary does not have to undergo physical image acquisition before classification is performed by a DCNN.  For example, this natural defensive mechanism of DCNNs does not apply to an adversary who is attempting to bypass an image filter or change the amount deposited into their own bank account, because the input to the DCNN is coming directly from an adversary without physical acquisition of the image.

\section*{Acknowledgment}

Supported in part by Research Experience for Undergraduates (REU) program at the Univ. of Colorado Colorado Springs (NSF Award No. 1359275) and in part by NSF IIS-1320956.

\bibliographystyle{IEEEtran}
\bibliography{main}

\end{document}